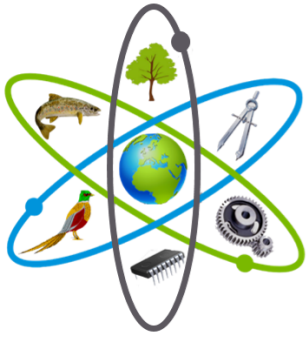

- RESEARCH ARTICLE -

# Generative Autoencoder Kernels on Deep Learning for Brain Activity Analysis

Gokhan Altan*, Yakup Kutlu

Iskenderun Technical University, Computer Engineering Department, Hatay, Turkey

**Abstract**

Deep Learning (DL) is a two-step classification model that consists feature learning, generating feature representations using unsupervised ways and the supervised learning stage at the last step of model using at least two hidden layers on the proposed structures by fully connected layers depending on of the artificial neural networks. The optimization of the predefined classification parameters for the supervised models eases reaching the global optimality with exact zero training error. The autoencoder (AE) models are the highly generalized ways of the unsupervised stages for the DL to define the output weights of the hidden neurons with various representations. As alternatively to the conventional Extreme Learning Machines (ELM) AE, Hessenberg decomposition-based ELM autoencoder (HessELM-AE) is a novel kernel to generate different presentations of the input data within the intended sizes of the models. The aim of the study is analyzing the performance of the novel Deep AE kernel for clinical availability on electroencephalogram (EEG) with stroke patients. The slow cortical potentials (SCP) training in stroke patients during eight neurofeedback sessions were analyzed using Hilbert-Huang Transform. The statistical features of different frequency modulations were fed into the Deep ELM model for generative AE kernels. The novel Deep ELM-AE kernels have discriminated the brain activity with high classification performances for positivity and negativity tasks in stroke patients.

**Keywords:** Deep Learning, SCP, Hilbert-Huang Transform, Autoencoder, Deep ELM

## Introduction

Deep Learning (DL) is the specified type of machine learning algorithms by handling artificial neural network basics with many hidden layers, various kernels and excessive neuron sizes. The DL has ability to perform computer vision (Krizhevsky, Sutskever, & Hinton, 2012), time-series analysis (Allahverdi, Altan, & Kutlu, 2018; Altan, Kutlu, & Allahverdi, 2016), modeling diagnostic applications (Allahverdi, Altan, & Kutlu, 2016; Gokhan Altan, Kutlu, Pekmezci, & Nural, 2018; Gokhan Altan, Kutlu, Pekmezci, & Yayık, 2018), natural language processing and more. Whereas the most famous DL algorithm including convolution neural networks, deep belief networks (DBN), deep reinforcement learning and deep neural network models are referred with the computer vision and digital image analysis problems, the researchers are focused on reducing the training time with kernel-based solutions. Altan et al. utilized DL algorithms on respiratory sounds to differentiate the smokers and respiratory diseases lung sound using second-order difference plot analysis on three dimensional chaos distribution (Altan, Kutlu, Pekmezci, & Nural, 2018). Kutlu et al. used DL algorithms on the recognition of the fish species using morphometric features on the fish mages. They extracted 12 landmark points including fins, mouth and tail points to calculate the morphometric features (Kutlu, Altan, Iscimen, Dogdu, & Turan, 2017).

---

* *Corresponding Author: Gökhan Altan, e-mail, gokhan.altan@iste.edu.tr*

An electroencephalogram (EEG) signal is commonly utilized to detect neurological cases of the patients. The EEG records the electrical activity of the brain using various types of sensors which are placed at head. The neurology discipline focuses on diagnostic applications by addressing the relationships of brain, nerve and muscle disorders. Analyzing the frequency-time features of the EEG signals has a great potential for computerized diagnosis of the disorders. The recording electrodes are placed to the specified points on the head with proper recording caps (Sanei & Chambers, 2007). The EEG recording process may vary according to the sessions and the channel number during the neurofeedback sessions. The electrodes directly contract the skin for getting full quality signals, and usually supported with electrically conductive liquid solutions. The possible electrode placements and the channel names are depicted in Figure 1. The EEG is the basic biomedical signal for human-computer interactions (Bosch, Mecklinger, & Friederici, 2001; Hinterberger et al., 2004). It was also utilized to appoint the brainwave changes for the visual and auditory stimulus (Devrim, Demiralp, Kurt, & Yücesir, 1999; Pham et al., 2005) and color and shape-based images (Kutlu, Yayık, Yildirim, & Yildirim, 2017). Detecting neurological disorders using time-series analysis was main support of the engineering on the EEG for explaining the meaning for stimuli resultants (B. Kotchoubey et al., 1996; Boris Kotchoubey et al., 2001; Ozdemir & Yildirim, 2014; Schneider et al., 1993; Siniatchkin, Kirsch, Kropp, Stephani, & Gerber, 2000). The analysis on EEG provides specialist-independent assessments for normal monitoring and abnormal managements on disabled and patient groups. Brain-base disorders including epilepsy, stroke, tumor, sleep disorders have a monitoring possibility using EEG related methods (Sanei & Chambers, 2007). The fluctuation, sudden changes and instantaneous responses on the EEG for the stimulus give deterministic features for the signal analysis techniques (Altan, Kutlu, & Allahverdi, 2016b). EEG features may help the early diagnosis and diagnosis and management of the neurological diseases (Ozdemir & Yildirim, 2014).

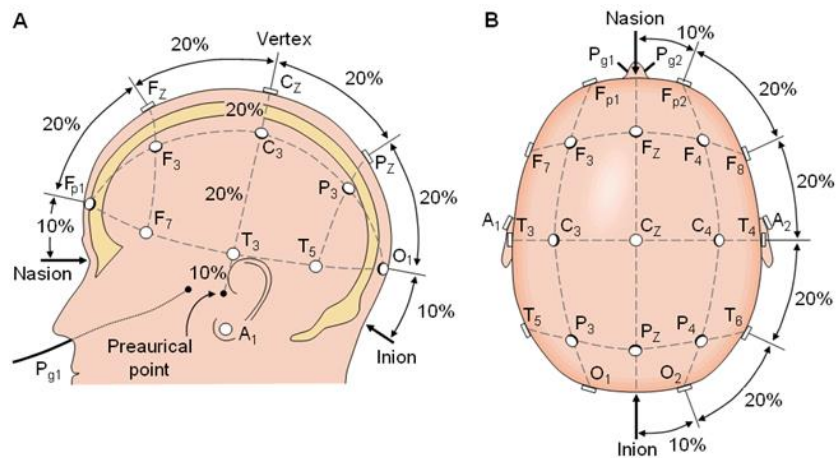

Fig. 1. EEG channels and electrode replacements on head (Malmivuo & Plonsey, 2012)

The EEG has different types including P300, Slow cortical potential (SCP), N400, and more. In this study the analysis was focused on SCP training signals. The SCP is non-invasive technique which has fractional changes at the cortical layer of the head (Ergenoglu et al., 1998). The duration of the SCP varies among 300ms and 10s (Birbaumer, Elbert, Canavan, & Rockstroh, 1990). The SCP is a low-frequency EEG component. The recording areas are the channels which were placed on the top center of the head during feedback sessions. In the consequence of the duration specifications of the SCP, it may contain readiness potentials, movement-related changes, P300 and N400 components (Stern, Ray, & Quigley, 2001). Negative SCP is related with depolarization of cortical neuronal cells and positive SCP is related with neuronal complication (Birbaumer et al., 1990). Ergenoglu et al. (1998) testified the relationship between SCP and P300 components. Devrim et al. (1999) proposed a model which stated the effect of the visual stimuli on the SCP. Kotchoubey et al. analyzed the SCP for detecting epilepsy for determining the influencing factors (B. Kotchoubey et al., 1996; Boris Kotchoubey et al., 2001). Siniatchkin et al. (2000) performed SCP analysis to detect the migraine on the signal changes. Bosch et al. (2001) detected the event-related potentials including object, spatial, and verbal information on the SCP training, Hinterberger et al. (2004) proposed a SCP-based tough translation device for subjects with amyotrophic lateral sclerosis. Göksu (2018) proposed an efficient brain-computer interface (BCI) by extracting wavelet transform features with neural network classifier model on SCP.

The remaining of this paper is organized as follows; the EEG database acquisition scenario, Hilbert-Huang Transform (HHT), the HHT-based statistical feature extraction processes, and the Deep Extreme Learning (Deep ELM) classifier are described in detail. The proposed SCP-based brain activity classifier model, Deep ELM kernels classification performances were compared with the advantages and disadvantages.

**Materials and Methods**

The database consists of SCP training of EEG recordings. The feature extraction step was performed using time-frequency-amplitude analysis by HHT. The HHT application and the statistical feature counting from the signal modulations were explained in this section. The conventional Deep ELM structure, ELM autoencoder (ELM-AE) and the novel kernel and Hessenberg decomposition-based ELM autoencoder (HessELM-AE) were enlightened.

### *Database*

The EEG is a diagnostic tool for potential brain problems. The EEG signals were acquired from two chronic stroke patients (Ruben, Helena, Andreas, & et al., 2014). There is one-week interval between the sessions for each patient. The EEG signals were recorded at a sampling rate of 256 Hz using a low-frequency EEG component from slow cortical potentials (SCP) training in stroke patients during eight neurofeedback sessions. Each neurofeedback session contains negativity and positivity trails. The trails have a duration of 8s which is comprised of baseline phase between 0-2s and active phase between 2s-8s. The Cz channel which is at the top of the head was selected for the analysis. During the neurofeedback sessions, the recorded brain activities were labeled as positivity and negativity for tasks successfully and unsuccessfully, respectively. Sessions 1-3 have 250 of positivity and negativity trials for each. Sessions 4-8 have 375 of negativity trials and 125 of positivity trials.

Table 1. Stimuli trial distribution of dataset

| Sessions | # Negativity Trials | # Positivity Trials |
|---|---|---|
| Session 1-4 | 2000 | 2000 |
| Session 5-8 | 3000 | 1000 |
| Total | 5000 | 3000 |

The number of the segmented trials are presented in Table 1. The trials have a non-homogeneous distribution in the dataset. It is important to have balanced training instances for each class in the machine learning approaches. Hence, same number of trials for positivity and negativity were utilized to train and the remaining trials were used to test the proposed classifier model.

### *Hilbert-Huang Transform*

The HHT is an analysis method for non-linear and non-stationary time series signals. The significant advantages of the HHT are adaptation ability and empirical theory. It has an extensive mathematical theory which has selective stoppage criteria (N. E. Huang & Wu, 2008; Oweis & Abdulhay, 2011). The HHT has been utilized in research areas for filtering, pre-processing, feature extraction and frequency spectral analysis for biomedical signal and more. Despite alternating stoppage criteria of the decomposition technique, mathematical theory is not completed yet (Hou & Tian, 2010). The HHT is a two-step transformation including Empirical Mode Decomposition (EMD) and Hilbert Transform (HT), sequentially. The EMD performs amplitude and time domain analysis by extracting Intrinsic Mode Functions (IMF). Each IMF is a different frequency modulation of the input time-series. The HT creates analytic function of instantaneous amplitude values from each IMF at the time-frequency domain (Hou & Tian, 2010; Ozdemir & Yildirim, 2014). The detailed formulations of the EMD and HT were presented in (Altan, Kutlu, & Allahverdi, 2016a).

*Deep Extreme Learning Machines*

The main advantage of the DL at the learning stage is defining the weights of hidden layer using unsupervised techniques for obtaining shared weights. Defining the pre-trained classification parameters provides attaining high generalization capability in conjunction with less optimization process at the supervised stage. One of the most open to improve the method among the unsupervised DL algorithms is autoencoder.

Autoencoders generate different presentations of the given input. The most known algorithms for DL are stacked denoising autoencoder and sparse autoencoder. The raising popularity of the DL has motivated the researchers into adapting new solutions for enhancing classification performances. ELM-AE is one of the most adaptable kernel for eliminating the over-fitting and reaching high generalization capability.

The main disadvantages of the DL are training time, big instance necessity, and over-fitting problem. The DL tries to overcome the excessive training time by predefined output weights and shared weights in the unsupervised stage of the algorithm. The ELM takes over with the specifications of generalization capability and fast training without iterations among the machine learning algorithms (Guang-bin Huang, Qin-yu Zhu, 2006). Whereas the integration of the ELM and DL will take advantages of both algorithms, the single hidden layered structure of the ELM has repressive factor for adapting the algorithm into the DL. The idea that the ELM can be adapted to the DL as autoencoders was only way transferring the capabilities of the ELM to the structures with many hidden layers. The ELM autoencoder was used to calculate the hidden layer weights by unsupervised ways. Adding supervised ELM classifier to the last layer with outputs creates Deep ELM model (Tang, Deng, & Huang, 2016).

The Deep ELM utilizes the conventional ELM-AE kernel. The ELM-AE calculates the output weights and neuron weights using Moore-Penrose matrix inverse basics (Barata & Hussein, 2012; Guang-bin Huang, Qin-yu Zhu, 2006). Altan et al. (2018) proposed a novel Deep ELM autoencoder kernel which focuses on Hessenberg decomposition-based inverse solutions. Altan et al. (2018) also proposed Lower-Upper triangulation based autoencoder (LUELM-AE) as an alternative to the existing kernels. They applied the Hessenberg decomposition based autoencoder (HessELM-AE) and LUELM-AE on respiratory sounds analysis and have achieved high classification performances for diagnosis of COPD.

**Experimental Results**

Deep ELM kernels have high generalization capabilities on various types of classification models. One of the main popularity of the Deep ELM models is easy adaptation to the machine learning algorithms for both supervised and unsupervised ways. Especially the BCI allows controlling the machines and automates using EEG signals. This is the strength side of the disabled people to facilitate daily works and duty-based processes. The BCI creates a connection with meaningful channels according to the processes. The SCP training is used to assess and detect the normal and abnormal patterns for disabled and neurological disorders. The aim of this study is evaluating the efficiency of the Deep AE kernels on brain activity classification for SCP training. The Stroke patients were selected for the neurological inability classification on positivity and negativity trials.

The HHT has a common use for many of non-stationary biomedical signals at the feature extraction stages. The EEG signals also were the focal point of the recent researches over last decade. Huang et al. (2008) implemented a BCI for determining the steady-state visual evoked potential using the HHT features. Li et al. (2009) focused on sleep stage classification on the EEG using the HHT as addition to the common processing approaches. They discussed on superiorities of the HHT about the efficiency and response time against wavelet transform. Ozdemir et al. and Oweis et al. utilized the HHT on EEG for detecting epileptic seizure. Ozdemir et al. (2014) achieved a classification accuracy rate of 89.66% and Oweis et al. (2011) has reached to classification accuracy rate of 94%. Taking into account of the capability of HHT on frequency-time decomposition, the HHT was utilized to extract significant frequency-time features on the SCP training from stroke patients. Figure 2 depicts the IMFs which were extracted from a random SCP training by EMD.

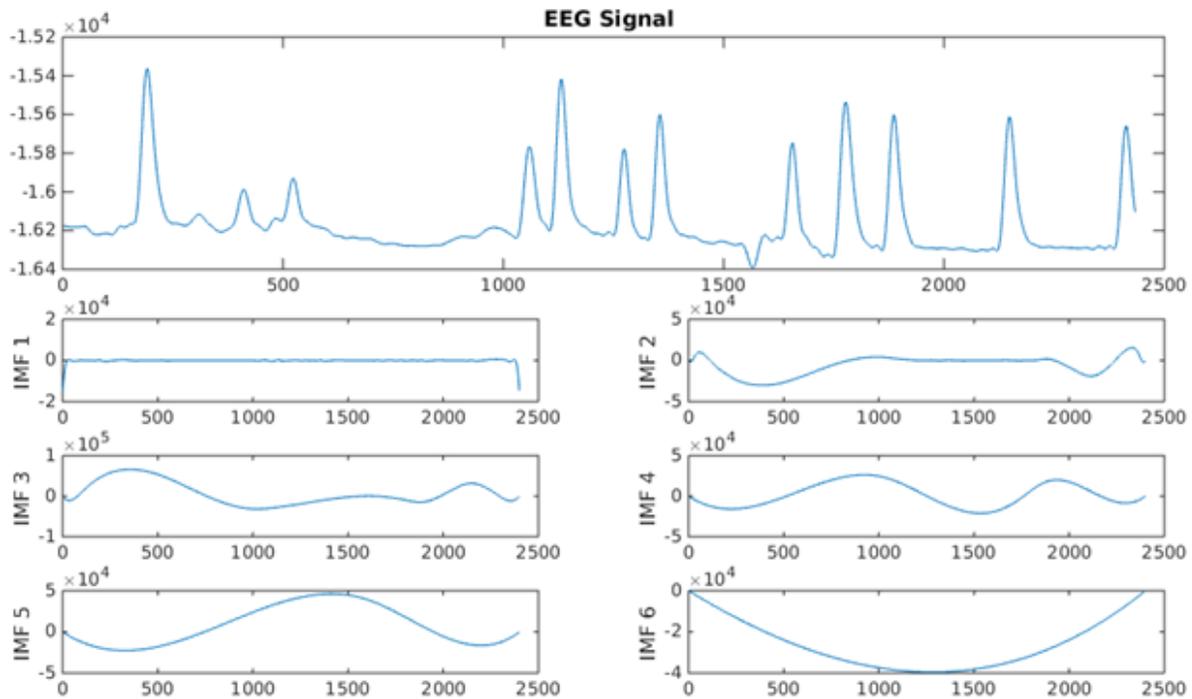

Figure. 2. A random EEG trial and extracted IMFs by EMD

The trials were segmented from the EEG database and the patient population was created. After implementing a low pass filter (10Hz) for reducing low frequency components each EEG for, Hilbert-Huang Transform was applied to the filtered EEGs. At the first stage EMD sifted the IMFs and the HT was applied to each IMF for obtaining time-frequency-energy distribution. The statistical features including standard deviation, minimum, maximum, correlation co-efficient, skewness, kurtosis, minimum, maximum, covariance, mode, moment, cumulant and mean were extracted from each IMF which represents different frequency modulations of the SCP training.

The patient population-based assessment is based on statistical valuation functions including accuracy, specificity, and sensitivity. The calculation of the performance metrics is performed using distributions which are correctly classified with positivity and negativity trails, and also the false classifications of the both trailer. The distribution of the classification is presented by contingency tables. The formulations of the classification metrics were detailed in (Altan, Kutlu, Pekmezci, & Nural, 2018).

The proposed Deep ELM models have negativity and positivity trials as outputs. The classification performances were evaluated using DL models with two and three hidden layers of which neuron size vary at a range of 100-500 increased by 10 neurons. The ELM-AE and HessELM-AE kernels were utilized for unsupervised definition of the output weights on the model. One of the biggest advantage of the Deep AE approach is excluding epochs and iterations during training. At the last stage of the Deep ELM model, the HessELM and ELM algorithms applied to the defined output parameters according to the outputs. Whereas the Deep ELM models were tested with a limited number of the parameters, the highest performance achievements were focused on. The classification performances were compared with (Altan et al., 2016) to assess the efficiency of the Deep ELM-AE kernels. The achievements for two and three hidden layers for DL approaches are presented in Table 2.

Table 2. Performances of the Deep Learning algorithms

| Models | | Classification Performances (%) | | |
|---|---|---|---|---|
| **Hidden Units** | **Classifier** | **Selectivity** | **Sensitivity** | **Accuracy** |
| 190-100 units | DBN | 94.75 | 87.60 | 77.85 |
| 350-100-260 units | DBN | 96.58 | 90.30 | 91.15 |
| 320-430 units | Deep ELM-AE | 94.04 | 85.60 | 85.37 |
| 440-210-220 units | Deep ELM-AE | 98.88 | 97.26 | 92.55 |
| 280-390 units | Deep HessELM-AE | 95.01 | 87.96 | 86.45 |
| 370-430-260 units | Deep HessELM-AE | 99.25 | 98.16 | 93.50 |

5-fold cross validation technique was utilized for the proposed models to incorporate each EEG trail for both training and the testing. In this validation technique the dataset was partitioned into 5 folds. Whereas four folds are used to train the classifier model, the remaining fold is disposed to test the model. The testing process continues until testing is completed for each fold. It is important to distribute the positivity and negativity trails in the training, homogeneously. The five testing results were averaged for determining overall performance (Wong, 2015).

The Deep ELM-AE has classified the trials with classification performance rates of 94.04%, 85.60%, and 85.37% for selectivity, sensitivity, and accuracy using two hidden layers' structure. The achievements are overtly high from the DBN classifier with two hidden layers. The HessELM-AE kernel also has achieved performance rates of 95.01%, 87.96%, and 86.45% for selectivity, sensitivity, and accuracy on the Deep ELM model with two hidden layers. The idea using increasing number of the hidden layers provides more detailed analysis of the frequency modulation features was realized for the both Deep ELM-AE and HessELM-AE kernels. The Deep ELM model with the conventional ELM-AE kernel has separated the positivity and negativity trials with classification performance rates of 98.88%, 97.26%, and 92.55% for selectivity, sensitivity, and accuracy. The Deep ELM model with HessELM-AE kernel has classified the brain activities with the rates of 99.25%, 98.16%, and 93.50% for selectivity, sensitivity, and accuracy.

**Discussion**

The SCP training databases have different characteristics in literature. That is why it is hard to compare the classification resultants with each other. The positivity and negativity trials are both focused on the SCP training for motor imaginary progresses. The reported analysis for the database has a significant separating achievements for two trials. Altan et al. utilized the HHT based statistical features on neural network and the DBN classifiers. They reported the superiority efficiency of DBN against neural networks. The ELM-AE kernels had been applied on the same feature set to specify the superiority and inferiority of the algorithms on the classification of the trials.

The same structures with the neuron size and hidden layer size were evaluated in the experiments. The same increment size by 10 neurons were iterated for both three the DL algorithms. The DBN achievements were directly contracted using the best achievements for the iterated output functions, learning rates, other classification parameters (Altan et al., 2016).

The Deep ELM-AE has achieved better classification performance than the DBN for brain activity classification. The ELM-AE has advantages of high generalization capability without iterations against the DBN. The DBN needs iterative unsupervised and supervised stages for training the model. But the Deep ELM-AE kernels do not need iterations in the defining unsupervised weight and optimization algorithms for high generalization capability. Considering three DL algorithms, the most accurate DL algorithm is HessELM-AE kernel for the HHT-based statistical features.

Whereas the Deep ELM-AE has discriminated the trials better performance than the DBN classifier for two hidden layers, the Deep ELM with HessELM kernel is the model with highest performance. Using more hidden layers supported enhancing the classification performance by deep analysis of statistical IMF features for each DL algorithm. The iterative supervised learning of the DBN by neural network models had extended the training time, even though the DBN had predefined classification parameters including output weights, biases and neuron size with unsupervised learning. The Deep ELM model with the conventional ELM-AE and HessELM-AE had advantage of being non-iterative method for training time. The Deep ELM model with HessELM-AE kernel has reported an accomplished classification for the brain activities with rates of 99.25%, 98.16%, and 93.50% for selectivity, sensitivity, and accuracy. It generalized the brain activities with the highest capacity among the aforementioned DL algorithms using the integration of HHT-based statistical model.

Due from using ELM-AE ensures defining the output weights without iteration for the supervised training, the supervised stage of the DL algorithm and according as the training of the proposed model require less time than the neural network included DL algorithms including DBN. Wherefore the EEG implementations are hard to extract the meaning depending upon instantaneous changes in the signals and noise, the BCIs need steady and accelerated classification models. The DL algorithms particularly the Deep ELM have handled the mentioned basic requirements of the BCI with the superiority on accuracy and simple solutions during supervised and unsupervised training.

The HHT is one of the most efficient way to detect the spectral and frequency analysis on time-series. Using HHT features for the trial classification of patients in stroke has performed significant resultants. Assigning the highest responsible frequency modulations for the SCP training sessions is important for designing and implementing BCIs. Supporting the distinctive feature extraction stages with influential machine learning algorithms enhances use of BCI for motor imaginary progresses. The integration of HHT features with the Deep ELM-AE kernel has increased the capability of the brain activity classification.

**Conclusion**

The novel Deep HessELM-AE kernel has discriminated the brain activity with high classification performances for positivity and negativity tasks in stroke patients. The capability of the HHT on the time series has handled the extracting significant characteristics of the different frequency modulation on spectral analysis.

The effectiveness of the DL algorithms has undisputed superiority over machine learning algorithms in literature. One of the useful specification of the DL is that it has feature learning stage and may not need feature extraction stages. In this study, we utilized the feature learning stage according to the frequency modulation characteristics instead of using the SCP training EEG signals directly. Hence, the EEG is a noisy and has instantaneously changes in milliseconds, the DL assessments on the time-series lead to learning the noisy data. And also using raw signal in the DL algorithms with feature learning stages for the big number of data points results in extending the training time for the models related with neuron sizes, iterations.

On the ground that the SCP is interrelated with brain related cognitive progresses, it has a common use for motor imaginary applications on the BCI. The integration of the EEG and computerized analysis supports defining the specific electrical patterns for cognitive trials. The developments on BCIs enable the people with the neurological disorders to control prosthetic limbs and objects to be controlled. Our study contains alternative kernels for enhancing the brain detection processes for the patients in stroke. Consequently, the achieved results indicate the efficiency of the DL algorithms on the HHT-based frequency modulation features for SCP training.

**References**

Allahverdi, N., Altan, G., & Kutlu, Y. (2016). Diagnosis of Coronary Artery Disease Using Deep Belief Networks. *2. International Conference on Engineering and Natural Science, Sarajevo, Bosnia*, 40–46.

Allahverdi, N., Altan, G., & Kutlu, Y. (2018). Deep Learning for COPD Analysis Using Lung Sounds. In *1st International Conference on Control and Optimization with Industrial Applications (COIA)* (pp. 74–76). Baku, Azerbaijan.

Altan, G., & Kutlu, Y. (2018). Hessenberg Elm Autoencoder Kernel For Deep Learning. *Journal of Engineering Technology and Applied Sciences*, *3*(2), 141–151. https://doi.org/10.30931/jetas.450252

Altan, G., Kutlu, Y., & Allahverdi, N. (2016a). A new approach to early diagnosis of congestive heart failure disease by using Hilbert–Huang transform. *Computer Methods and Programs in Biomedicine*, *137*, 23–34. https://doi.org/10.1016/J.CMPB.2016.09.003

Altan, G., Kutlu, Y., & Allahverdi, N. (2016b). Deep Belief Networks Based Brain Activity Classification Using EEG from Slow Cortical Potentials in Stroke. *International Journal of Applied Mathematics, Electronics and Computers*, *4*(Special Issue-1), 205–210. https://doi.org/10.18100/ijamec.270307

Altan, G., Kutlu, Y., Pekmezci, A. Ö., & Nural, S. (2018). Deep learning with 3D-second order difference plot on respiratory sounds. *Biomedical Signal Processing and Control*, *45*, 58–69. https://doi.org/10.1016/j.bspc.2018.05.014

Altan, G., Kutlu, Y., Pekmezci, A. Ö., & Nural, S. (2018). Deep learning with 3D-second order difference plot on respiratory sounds. *Biomedical Signal Processing and Control*, *45*, 58–69. https://doi.org/10.1016/j.bspc.2018.05.014

Altan, G., Kutlu, Y., Pekmezci, A. Ö., & Yayık, A. (2018). Diagnosis of Chronic Obstructive Pulmonary Disease using Deep Extreme Learning Machines with LU Autoencoder Kernel. In *7th International Conference on Advanced Technologies (ICAT'18)* (pp. 618–622). Antalya.

Barata, J. C. A., & Hussein, M. S. (2012). The Moore-Penrose Pseudoinverse: A Tutorial Review of the Theory. *Brazilian Journal of Physics*. https://doi.org/10.1007/s13538-011-0052-z

Birbaumer, N., Elbert, T., Canavan, A. G., & Rockstroh, B. (1990). Slow potentials of the cerebral cortex and behavior. *Physiological Reviews*, *70*(1), 1–41.

Bosch, V., Mecklinger, A., & Friederici, A. D. (2001). Slow cortical potentials during retention of object, spatial, and verbal information. *Cognitive Brain Research*, *10*(3), 219–237. https://doi.org/10.1016/S0926-6410(00)00040-9

Devrim, M., Demiralp, T., Kurt, A., & Yücesir, I. (1999). Slow cortical potential shifts modulate the sensory threshold in human visual system. *Neuroscience Letters*, *270*(1), 17–20. https://doi.org/10.1016/S0304-3940(99)00456-5

Ergenoglu, T., Demiralp, T., Beydagi, H., Karamürsel, S., Devrim, M., & Ermutlu, N. (1998). Slow cortical potential shifts modulate P300 amplitude and topography in humans. *Neuroscience Letters*, *251*(1), 61–64. https://doi.org/10.1016/S0304-3940(98)00498-4

Göksu, H. (2018). BCI oriented EEG analysis using log energy entropy of wavelet packets. *Biomedical Signal Processing and Control*, *44*, 101–109. https://doi.org/10.1016/j.bspc.2018.04.002

Guang-bin Huang, Qin-yu Zhu, C. S. (2006). Extreme learning machine: A new learning scheme of feedforward neural networks. *Neurocomputing*. https://doi.org/10.1109/IJCNN.2004.1380068

Hinterberger, T., Schmidt, S., Neumann, N., Mellinger, J., Blankertz, B., Curio, G., & Birbaumer, N. (2004). Brain-computer communication and slow cortical potentials. *IEEE Transactions on Biomedical Engineering*, *51*(6), 1011–1018. https://doi.org/10.1109/TBME.2004.827067

Hou, Y., & Tian, H. (2010). An automatic modulation recognition algorithm based on HHT and SVD. In *Proceedings - 2010 3rd International Congress on Image and Signal Processing, CISP 2010* (Vol. 8, pp. 3577–3581). https://doi.org/10.1109/CISP.2010.5647536

Huang, M., Wu, P., Liu, Y., Bi, L., & Chen, H. (2008). Application and contrast in brain-computer interface Between hilbert-huang transform and wavelet transform. In *Proceedings of the 9th International Conference for Young Computer Scientists, ICYCS 2008* (pp. 1706–1710). https://doi.org/10.1109/ICYCS.2008.537

Huang, N. E., & Wu, Z. (2008). a Review on Hilbert-Huang Transform: Method and Its Applications. *October*, *46*(2007), 1–23. https://doi.org/10.1029/2007RG000228.1.INTRODUCTION

Kotchoubey, B., Schneider, D., Schleichert, H., Strehl, U., Uhlmann, C., Blankenhorn, V., … Birbaumer, N. (1996). Self-regulation of slow cortical potentials in epilepsy: A retrial with analysis of influencing factors. *Epilepsy Research*, *25*(3), 269–276. https://doi.org/10.1016/S0920-1211(96)00082-4

Kotchoubey, B., Strehl, U., Uhlmann, C., Holzapfel, S., König, M., Fröscher, W., … Birbaumer, N. (2001). Modification of slow cortical potentials in patients with refractory epilepsy: a controlled outcome study. *Epilepsia*, *42*(3), 406–416.

Krizhevsky, A., Sutskever, I., & Hinton, G. E. (2012). ImageNet Classification with Deep Convolutional Neural Networks. *Advances In Neural Information Processing Systems*. https://doi.org/http://dx.doi.org/10.1016/j.protcy.2014.09.007

Kutlu, Y., Altan, G., Iscimen, B., Dogdu, S. A., & Turan, C. (2017). Recognition of Species of Triglidae Family using Deep Learning. *Journal of Black Sea / Mediterranean Environment*, *23*(1), 56–65. Retrieved from http://www.blackmeditjournal.org/index.php/component/k2/item/574

Kutlu, Y., Yayık, A., Yildirim, E., & Yildirim, S. (2017). LU triangularization extreme learning machine in EEG cognitive task classification. *Neural Computing and Applications*, pp. 1–10. https://doi.org/10.1007/s00521-017-3142-1

Li, Y., Yingle, F., Gu, L., & Qinye, T. (2009). Sleep stage classification based on EEG hilbert-huang transform. In *2009 4th IEEE Conference on Industrial Electronics and Applications, ICIEA 2009* (pp. 3676–3681). https://doi.org/10.1109/ICIEA.2009.5138842

Malmivuo, J., & Plonsey, R. (2012). *Bioelectromagnetism: Principles and Applications of Bioelectric and Biomagnetic Fields*. *Bioelectromagnetism: Principles and Applications of Bioelectric and Biomagnetic Fields*. https://doi.org/10.1093/acprof:oso/9780195058239.001.0001

Oweis, R. J., & Abdulhay, E. W. (2011). Seizure classification in EEG signals utilizing Hilbert-Huang transform. *Biomedical Engineering Online*, *10*, 38. https://doi.org/10.1186/1475-925X-10-38

Ozdemir, N., & Yildirim, E. (2014). Patient specific seizure prediction system using hilbert spectrum and Bayesian networks classifiers. *Computational and Mathematical Methods in Medicine*, *2014*. https://doi.org/10.1155/2014/572082

Pham, M., Hinterberger, T., Neumann, N., Kübler, A., Hofmayer, N., Grether, A., … Birbaumer, N. (2005). An auditory brain-computer interface based on the self-regulation of slow cortical potentials. *Neurorehabilitation and Neural Repair*, *19*(3), 206–218. https://doi.org/10.1177/1545968305277628

Ruben, R., Helena, E., Andreas, H., & et al. (2014). *Slow cortical potential training in stroke*. Germany.

Sanei, S., & Chambers, J. a. (2007). *EEG Signal Processing*. *Chemistry & biodiversity* (Vol. 1). https://doi.org/10.1002/9780470511923

Schneider, F., Elbert, T., Heimann, H., Welker, a, Stetter, F., Mattes, R., … Mann, K. (1993). Self-regulation of slow cortical potentials in psychiatric patients: alcohol dependency. *Biofeedback and Self-Regulation*, *18*(1), 23–32.

Siniatchkin, M., Kirsch, E., Kropp, P., Stephani, U., & Gerber, W. D. (2000). Slow cortical potentials in migraine families. *Cephalalgia*, *20*(10), 881–892. https://doi.org/10.1046/j.1468-2982.2000.00132.x

Stern, R. M., Ray, W. J., & Quigley, K. S. (2001). Psychophysiological recording (2nd ed.). *Journal of Psychophysiology*. https://doi.org/10.1027//0269-8803.15.1.47

Tang, J., Deng, C., & Huang, G.-B. (2016). Extreme Learning Machine for Multilayer Perceptron. *IEEE Transactions on Neural Networks and Learning Systems*, *27*(4), 809–821. https://doi.org/10.1109/TNNLS.2015.2424995

Wong, T.-T. (2015). Performance evaluation of classification algorithms by k-fold and leave-one-out cross validation. *Pattern Recognition*, *48*(9), 2839–2846. https://doi.org/http://dx.doi.org/10.1016/j.patcog.2015.03.009